\documentclass{article}

\usepackage{microtype}
\usepackage{graphicx}
\usepackage{subfigure}
\usepackage{booktabs} 

\usepackage{hyperref}

\usepackage[accepted]{icml2023}

\usepackage{amsmath}
\usepackage{amssymb}
\usepackage{mathtools}
\usepackage{amsthm}

\usepackage[capitalize,noabbrev]{cleveref}

\theoremstyle{plain}

\theoremstyle{definition}

\theoremstyle{remark}

\usepackage[textsize=tiny]{todonotes}

\usepackage{subfigure}
\newcommand{\ie}{\emph{i.e.,}\xspace}
\newcommand{\eg}{\emph{e.g.,}\xspace}

\usepackage{graphicx}
\usepackage{caption}
\usepackage{bm}
\usepackage{booktabs}
\usepackage{xspace}
\usepackage{color}
\usepackage{caption}
\usepackage{wrapfig}
\usepackage{graphicx}

\icmltitlerunning{A Study on Transformer Configuration and Training Objective}

\begin{document}

\twocolumn[
\icmltitle{A Study on Transformer Configuration and Training Objective}




\begin{icmlauthorlist}
\icmlauthor{Fuzhao Xue}{nus}
\icmlauthor{Jianghai Chen}{nus}
\icmlauthor{Aixin Sun}{ntu}
\icmlauthor{Xiaozhe Ren}{hw}
\icmlauthor{Zangwei Zheng}{ntu}
\icmlauthor{Xiaoxin He}{nus}
\icmlauthor{Yongming Chen}{ntueee}
\icmlauthor{Xin Jiang}{hw}
\icmlauthor{Yang You}{nus}
\end{icmlauthorlist}

\icmlaffiliation{nus}{School of Computing, National University of Singapore}
\icmlaffiliation{ntu}{School of Computer Science and Engineering, Nanyang Technological University}
\icmlaffiliation{ntueee}{School of Electrical and Electronic Engineering, Nanyang Technological University}
\icmlaffiliation{hw}{Huawei Noah’s Ark Lab}

\icmlcorrespondingauthor{Fuzhao Xue}{f.xue@u.nus.edu}

\icmlkeywords{Machine Learning, ICML}

\vskip 0.3in
]



\printAffiliationsAndNotice{}  

\begin{abstract}
Transformer-based models have delivered impressive results on many tasks, particularly vision and language tasks. In many model training situations, conventional configurations are often adopted. For example, we usually set the base model with hidden size (\ie model width) to be 768 and the number of transformer layers (\ie model depth) to be 12. In this paper, we revisit these conventional configurations by studying the the relationship between transformer configuration and training objective. We show that the optimal transformer configuration is closely related to the training objective. Specifically, compared with the simple classification objective, the masked autoencoder is effective in alleviating the over-smoothing issue in deep transformer training. Based on this finding, we propose ``Bamboo'', a notion of using deeper and narrower transformer configurations, for masked autoencoder training. On ImageNet, with such a simple change in configuration, the re-designed Base-level transformer achieves 84.2\% top-1 accuracy and outperforms SoTA models like MAE by $0.9\%$. On language tasks, re-designed model outperforms BERT with the default setting by 1.1 points on average, on GLUE benchmark with 8 datasets.

\end{abstract}

\section{Introduction}\label{Introduction}

Transformer-based language models have achieved promising results on natural language understanding tasks, \eg  Q\&A~\citep{qu2019bert,yang2020bert}, relation extraction~\citep{xue2020embarrassingly, zhou2020document} and dialogue system~\citep{ni2021recent}. Recently, on vision tasks, transformers~\citep{dosovitskiy2020image,zhou2021deepvit, xue2021go, xue2022one} also outperform convolution-based models by a large margin. With sufficient training data,  transformer-based models can be scaled to trillions of trainable parameters~\citep{fedus2021switch, du2021glam}. Through scaling along the width (\ie hidden dimension) and depth (\ie number of transformer blocks), these huge transformers show effectiveness across various tasks and even areas.


\textbf{Where are the configurations from?} When using transformer, we typically follow the existing work to set the same width and depth for a ``fair'' comparison. For instance, we usually set the width of transformer-base model as 768 and the depth as 12. An interesting question here is: \textit{Why do we select these hyper-parameters, even for problems in different areas?} To answer this question, we revisit the conventional configurations from some representative studies. For vision transformer~\citep{dosovitskiy2020image}, authors set the base ViT configuration according to those used in BERT~\citep{devlin2018bert}.  BERT selects such configuration following OpenAI GPT~\citep{radford2018improving}. OpenAI also follows the original transformer paper~\citep{vaswani2017attention}. In the original transformer paper, \citet{vaswani2017attention} conduct a set of ablation studies on machine translation task to find the optimal configurations. That is, for a good range of tasks, we have largely followed the transformer configuration based on an ablation study on machine translation task, \ie a sequence-to-sequence task.


\textbf{Should we use the same configuration for different training objectives?} Nowadays, transformer-based models can be trained with various training objectives or strategies~\citep{tay2022unifying,tay2022transcending}. Taking the vision transformer~\citep{dosovitskiy2020image} as an example, we can train transformer from scratch with a supervised learning setting for image classification. 
In this straightforward image classification task, each image is modeled as a sequence of  tokens, and each token corresponds to a patch in the image. We use the  global information (from all tokens/patches of the image) to predict a single label, the category of the image. Here, as the training objective is to capture the global information of an image, the differences between token representations would not be considered directly. This image classification task is quite different from  machine translation task, which requests for a strong understanding of a token sequence and generating another sequence. Hence, intuitively, it is natural to assume that different optimal transformer configurations exist for these two different tasks. 

\textbf{Over-smoothing issue of the simple classification training objective.} Previous work has tried to train a deeper transformer from scratch. However, as reported in~\citep{zhou2021deepvit,gong2021vision}, training by classification task (\ie using the global signal of the input sequence) has the over-smoothing problem. That means, at the deeper transformer layers, all token representations tend to be identical~\citep{Brunner2020On}. Such issue harms the scalability of training vision transformer, especially the scaling along depth. When scaling to a larger model, we only get a slight improvement or even poorer accuracy. Recently, \citet{zhou2021deepvit, gong2021vision} show that, when adding special-designed regularization to avoid the ``uniform tokens'' (\ie the over-smoothing problem), it is possible to train a deeper transformer on the sequence (image) classification setting.

\textbf{Masked autoencoder can scale to deeper and wider models without additional training data.} Different from training from scratch above, the masked autoencoder is a two-stage training framework, including pre-training and fine-tuning. Given a partially masked input sequence, the pre-training stage aims to recover the original unmasked sequence. The fine-tuning is similar to the aforementioned training from scratch but requires much fewer training epochs. 
With the masked autoencoder, recent studies~\citep{bao2021beit,he2021masked} successfully train large-scale transformers, even without using additional training data compared to supervised learning. This is counterintuitive because we usually assume more available training data is the key of self-supervised learning to improve effectiveness. 
This result motivates us to rethink the reason behind it.

\textbf{Masked autoencoder can alleviate the over-smoothing issue.} Intuitively, in masked autoencoder frameworks (\eg BERT, BEiT), the target is to recover the masked tokens based on the unmasked tokens. Compared to training transformer from scratch under supervised learning, whose target is a simple classification task, the masked autoencoder framework adopts a sequence labeling target.
We hypothesize that the masked autoencoder training can alleviate the over-smoothing issue, which is a possible reason why the masked autoencoder can help to scale transformer up. Specifically, the sequence labeling task requires the model to learn semantic information from neighboring unmasked tokens. Since different masked tokens have different unmasked neighboring tokens, the unmasked token representations must carry their corresponding and sufficient semantics for the accurate prediction of the masked tokens, which in turn prevents the token representations to become identical (or very similar to each other).
In a word, we may infer that the masked autoencoder's training objective helps to alleviate the over-smoothing problem by its regularization on token differences. To justify the reasoning above, we conduct an experimental investigation in Section~\ref{sec: experimental investigation}. 
The results show that the over-smoothing issue is indeed alleviated in the masked autoencoder. Compared to training under masked autoencoder, training transformer by a simple classification task (\eg training vision transformer from scratch) does not have such benefit.

\textbf{Why and how masked autoencoder alleviates over-smoothing?} We further explore the reason behind this phenomenon via Fourier domain analysis in Section~\ref{sec: theoretical reasoning}. First, self-attention layer in transformer will decay the high-frequency component of input signal~\citep{wang2022anti}. When all high-frequency components are erased, all token representations would be identical. We find the masked autoencoder training objective can be seen as reconstructing the high-frequency components (HC) of input signal from the HC of the noisy masked input signal. Therefore, masked autoencoder can alleviate over-smoothing via learning a slower HC decay rate. Such ability is achieved by training the weights in self-attention layer. To further verify this finding, we conduct quantitative analysis in Section~\ref{sec: quantitative verification} and results show that, compared with the model trained with simple classification objective, the trainable matrices in model trained with masked autoencoder objective indeed has slower HC decay.

\textbf{Potential of masked autoencoder with deeper configurations.} If the masked autoencoder alleviates the over-smoothing issue (which is a challenge for scaling transformer along depth), does this mean the masked autoencoder can get more benefits from deep configurations? To answer this question, we re-visit the configurations for different training objectives, especially for the masked autoencoder. Accordingly, we conduct experiments to investigate the masked autoencoder configurations and propose our idea, Bamboo\footnote{The narrower and taller shape of the re-designed transformer looks like bamboo.}. When training transformer with masked autoencoder, we suggest using deeper and narrower configurations with comparable computation budget as a typical setting, to achieve better effectiveness. To evaluate our new configurations, we conduct comprehensive experiments on computer vision and natural language processing tasks. On vision tasks, we evaluate our configuration on large-scale vision transformer training. With Bamboo configuration, the masked autoencoder outperforms baseline by a large margin. For instance, on ImageNet, with a comparable number of trainable parameters and computational cost, our narrower and deeper base-scale masked autoencoder, Bamboo-B, outperforms MAE-B by 0.9\% in terms of top-1 accuracy. On natural language processing tasks, we conduct experiments on BERT. Results show that our configurations can improve BERT-L by 1.1 points on GLUE datasets.

\textbf{Contributions} In summary, our main contributions are three folds: 1) We first study the relationship between transformer configuration and training objective, and then propose the insight that the masked autoencoder helps transformer to handle over-smoothing, although there can be other cofounders like training stability. We show this finding by experimental investigation, and more importantly, by theoretical reasoning on Fourier domain and verify our reasoning via quantitative analysis;
2) We argue that the existing transformer configurations cannot fully use the strength of masked autoencoder. To this end, we propose Bamboo, an idea to scale transformer along depth when training with masked autoencoder. We show that the narrower and deeper versions overperform  existing configurations, in a plug-and-play manner; 
3) We further verify our Bamboo configurations on larger scale vision transformer pre-training and natural language tasks. Results show that our Bamboo achieves state-of-the-art top-1 accuracy on image classification, and outperforms the original BERT configurations by 1.1 points on GLUE.




\textbf{TL;DR for Practitioners} In this paper, the most important thing we want to highlight is, not to underestimate the training objective before tuning model configuration. Usually, for a fair comparison, we simply adopt the previous configurations. However, sometimes, one training objective may look decent if it wins the ``configuration lottery''\footnote{The previous used configuration matches well with the new training objective.}. However, for a different objective, the effectiveness would be underestimated without a configuration sweep. We may then miss a good training objective for the community. Therefore, to know about the potential of each novel training objective design, we strongly suggest practitioners analyze the inductive bias and customize configurations. Our paper shows one example of such analysis on MAE.

The following sections are organized based on the analysis process of this work. In Section~\ref{over-smoothing in Masked Autoencoder}, we briefly review the over-smoothing problem in transformer and show the strength of masked autoencoder in handling this issue. We then conduct experiments to investigate scaling masked autoencoder along the depth in Section~\ref{Bamboo}. Based on the consistent sweet depth across scales, we suggest the Bamboo idea, using narrower and deeper configurations in masked autoencoder training. Then, we adapt the new configurations to a larger scale, and conduct evaluations across different areas, vision tasks in Section~\ref{Evaluation Vision} and NLP tasks in Section~\ref{Evaluation Language}. Finally, we discuss the difference between this work and the related work in Section~\ref{Discussion}.

\section{Over-smoothing under Different Training Objectives}\label{over-smoothing in Masked Autoencoder}

The over-smoothing issue is well noted in graph neural networks. When we stack many graph convolution networks, the node representations tend to be identical~\citep{chen2020measuring}. Recent studies show that transformer has a similar problem, known as ``uniform tokens''~\citep{shi2021revisiting}. In deep transformer, each token representation can be seen as a node in a graph, and each attention score is an edge. Different token representations tend to be identical when we scale transformer across depth. In this work, we mainly focus on the over-smoothing problem under different training objectives, \ie sequence-level supervised learning, and masked antoencoder-based self-supervised learning. To compare these two objectives, we use supervised vision transformer (ViT)~\citep{dosovitskiy2020image} and vision masked autoencoder (MAE)~\citep{he2021masked} as two representative platforms to show our insights.

\subsection{Experimental Investigation}\label{sec: experimental investigation}

\begin{figure}[t]

\centering
\subfigure[Average standard deviation]{\label{fig:over-smoothing-std}\includegraphics[width=0.45\textwidth]{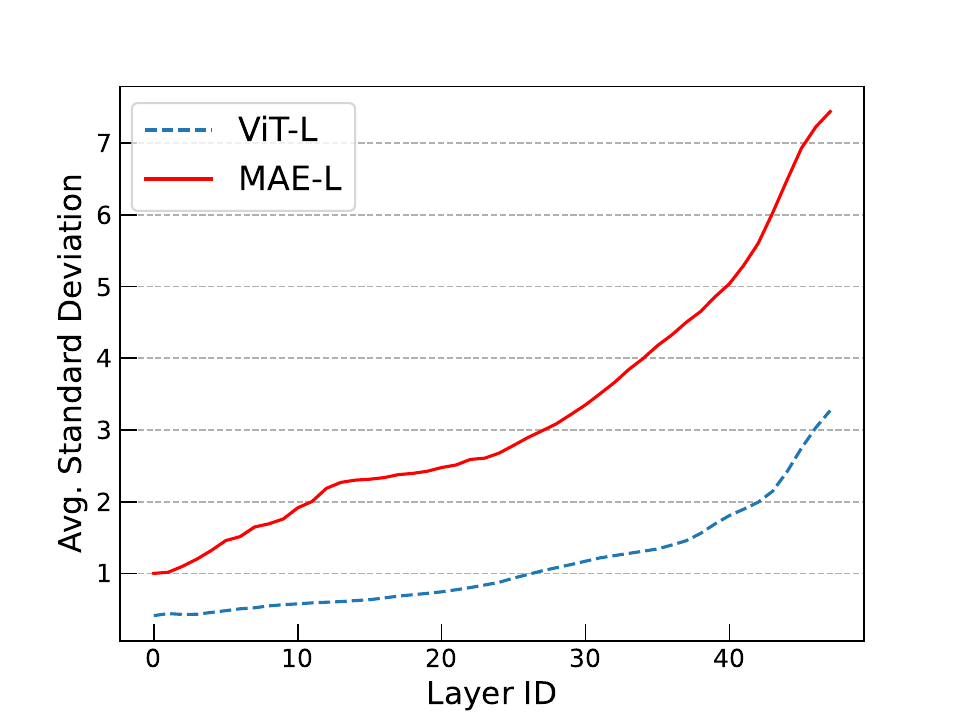}}\vspace{-0.4cm}
\subfigure[Average patch-pair cosine similarity]{\label{fig:over-smoothing-cos}\includegraphics[width=0.45\textwidth]{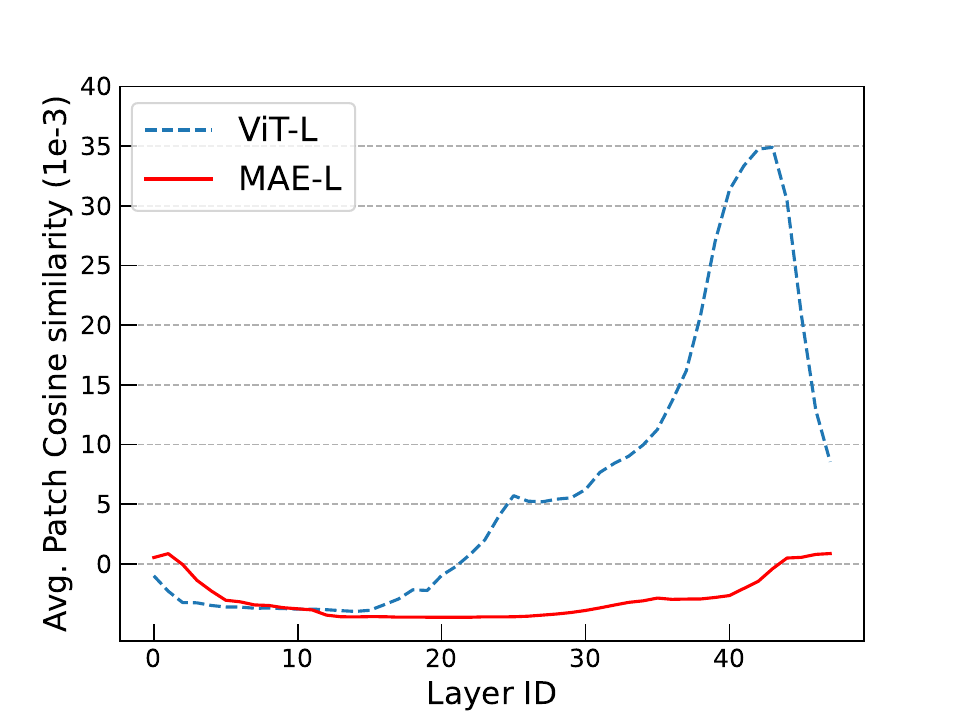}}\vspace{-0.35cm}
\caption{Over-smoothing analysis of ViT and MAE  with the same configuration (48 layers). We calculate the average standard deviation and average patch cosine similarity at every layer to compare the over-smoothing level.}
\label{fig:over-smoothing}
\vspace{-0.4cm}
\end{figure}

\textbf{Standard deviation.} We first conduct two sets of quantitative analysis to study the over-smoothing behaviour under the two training objectives. For the first set of quantitative analysis, we focus on the standard deviation of token representations at different transformer blocks. Formally, transformer block's output (\ie token representations) can be written as $h=\{h_0, h_1, ..., h_{T-1}\}$, where $h_t \in \mathbb{R}^{d}$, $d$ is the hidden dimension, $T$ is the number of tokens at different transformer blocks. Usually, $T$ is fixed across transformer blocks. For the token representations after the $l^{\mathrm{th}}$ transformer block, we denote the token representations as $h^l$. To compare the over-smoothing issue between supervised learning and masked autoencoder, we calculate the mean standard deviation of the token representations at different transformer layers:
\begin{equation}
    {ms}^{i} =  \sqrt{\frac{1}{T-1} \sum\limits_{t=0}^{T-1}  \left(h_t^i - \bar{h^i}\right)^2 }
\end{equation}
where $\bar{h^i}=\frac{1}{T-1}{\sum\limits_{t=0}^{T-1} h_t^i}$ is the mean vector of token representations at the $i^{\mathrm{th}}$ transformer block. 
We use a deeper version (48 layers) of ViT-L and MAE-L as platform\footnote{The default configuration of ViT-L and MAE-L includes only 24 layers. We use deeper configuration to expose the over-smoothing issue more clearly.} to validate our analysis above. Specifically, the ViT-L is trained on the supervised learning setting and treats image classification as a simple classification task, by average pooling prediction head. The MAE-L is trained on the masked autoencoder setting, which means the pre-training target is close to a sequence labeling task. We then fine-tune the pre-trained model for image classification. We set the width as 768 and the depth as 48. The total number of trainable parameters and FLOPs is comparable to the original configuration (\ie width is 1024, depth is 24). The the mean standard deviation of the token representations is shown in Figure~\ref{fig:over-smoothing-std}. 

Observe that the mean standard deviations of ViT-L and MAE-L both increase along depth. Intuitively, this contradicts with our expectation as the over-smoothing leads to similar token representations. As stated in~\citep{dong2021attention}, residual connection within transformer indeed alleviates over-smoothing (to some extent). Nevertheless, the over-smoothing issue still exists even if we have residual connection~\citep{dong2021attention,shi2021revisiting}.  We can observe that the mean standard deviation increasing on MAE-L is much faster than that on ViT-L. That means the deeper transformer blocks can learn different semantics for different token representations in MAE. In other words, the over-smoothing issue is much less pronounced in the transformer model pre-trained with the masked autoencoder setting.

\textbf{Patch-pair cosine similarity.} We further verify that the over-smoothing issue can be alleviated in masked autoencoder. Following \citet{gong2021vision}, we compare the patch-pair cosine similarity between ViT-L and MAE-L with deeper and narrower configuration. If there is an over-smoothing issue in the model, we should observe that the patch-pair cosine similarity increases along depth. The more serious the over-smoothing issue is, the faster the cosine similarity increases. To remove the impact from input representations (residual connection), we use the zero-centered token representations $\tilde{h^i}= h^i - \bar{h^i}$ for evaluation. The results are shown in Figure~\ref{fig:over-smoothing-cos}. Generally, the cosine similarity of ViT increases along the depth due to over-smoothing. However, for the model pre-trained by the masked autoencoder framework, the cosine similarity keeps constant along depth. This comparison is interesting, as we can barely observe the over-smoothing issue on the model pre-trained by the masked autoencoder, even if we are using a deeper model than usual.



\subsection{Theoretical Analysis}\label{sec: theoretical reasoning}

The reason why over-smoothing happens in Transformer has been well-studied~\citep{dong2021attention, wang2022anti}. Conceptually, each token representation can be seen as node in a directed graph and each attention score is a weighted edge. Recent study~\citep{wang2022anti} proposes to understand the transformer over-smoothing issue via the Fourier domain analysis by giving a closer examination of model architecture. However, existing work ignores that training objective also has relation with over-smoothing. This paper adapts their theorem as basis to reason why MAE can alleviate over-smoothing.

Given a Discrete Fourier Transform (DFT) $\mathcal{F}$: $\mathbb{R}^N \to \mathbb{C}^N $, the Inverse Discrete Fourier Transform (IDFT) $\mathcal{F}^{-1}$: $\mathbb{C}^N \to \mathbb{R}^N  $, and input signal $\mathbf{x} \in \mathcal{R}^N$, let $\mathbf{z}=\mathcal{F}\mathbf{x}$ be the spectrum of $x$.
$\mathbf{z}_{\mathcal{DC}}\in\mathbb{C}$ and $\mathbf{z}_{\mathcal{HC}}\in\mathbb{C}^{N-1}$ take the first element and the rest elements of $\mathbf{z}$, respectively. 
The DFT here can be implemented by left multiplying a pre-defined DFT matrix whose $k^\mathrm{th}$ row is Fourier basis $\mathnormal{f}_k=[e^{2\pi j(k-1)\cdot0},\dots,e^{2\pi j(k-1)\cdot(N-1)}]^T/\sqrt{N}$, where $j$ denotes the imaginary unit and $k$ denotes the $k$-th row of DFT matrix. 
Therefore, for signal $\mathbf{x}$, we have the Direct-Current (DC) component $\mathcal{DC}[\mathbf{x}]=\mathnormal{f}_1\mathbf{x}$ and the complementary high-frequency component $\mathcal{HC}[\mathbf{x}]=[\mathnormal{f}_2,\dots,\mathnormal{f}_N]\mathbf{x}$.

Based on the definition above, given input token sequence $\mathbf{X}\in \mathcal{R}^{N \times d} $, its $i$-th channel $\mathbf{x}_i \in \mathcal{R}^N$, and radius of a ball $\gamma > 0$, assuming $||\mathbf{x}_i||_2 \le \gamma^2$, \citet{wang2022anti} proposes and proves:

\begin{equation}\label{eq: smoothing rate}
    ||\mathcal{HC}[\mathrm{SA}(\mathbf{X})]||_{F} \le \tau ||\mathcal{HC}[\mathbf{X}]||_{F}
\end{equation}

\begin{equation}\label{eq: tau define}
    \tau =  \sqrt{\frac{ne^{2\alpha}}{e^{2\alpha}+n-1}} ||\mathbf{W}_V||_2
\end{equation}
\begin{equation}\label{eq: alpha define}
    \alpha \le \frac{\gamma^2||\mathbf{W}_Q{\mathbf{W}_K}^T||_2}{\sqrt{d}}
\end{equation}
where $\mathrm{SA}(\cdot)$ denotes self-attention~\citep{vaswani2017attention}, $\mathbf{W}_Q$, $\mathbf{W}_K$, $\mathbf{W}_V$ are query, key and value trainable matrices. When $\tau < 1$, $\mathcal{HC}[\mathrm{SA}(\mathbf{X})]$ will decay to zero exponentially. As a comparison, the DC component would not be affected by the attention scores. After applying a number of attention layers, which is exactly what we do in Transformer, the DC component would dominate the hidden representations and thus over-smoothing happens.

Then, our first question is, can we understand the reason why MAE can alleviate the over-smoothing with Fourier analysis? For MAE, we mask parts of the ground truth $\mathbf{X}$ as $\mathbf{X}^m$. We define the mask function as $\mathbf{X}^m=\mathcal{M}(\mathbf{X})$. The masked parts in $\mathbf{X}^m$ are filled with the DC Component as what we usually do in popular MAE (\eg BERT). We usually use $\mathbf{X}^m$ as the model input. During training, model is minimizing the distance between $\mathbf{X}^m_L$ and $\mathbf{X}$, where $\mathbf{X}^m_L$ is the hidden representation after $L$ transformer layers. If we transform $\mathbf{X}$, $\mathbf{X}^m$ and $\mathbf{X}^m_L$ to Fourier domain like $\mathbf{Z}=\mathcal{F}\mathbf{X}$, $\mathbf{Z}^m=\mathcal{F}\mathbf{X}^m$, and $\mathbf{Z}^m_L=\mathcal{F}\mathbf{X}^m_L$, the learning objective is minimizing the distance between $||\mathcal{HC}[\mathbf{Z}]||_F$ and $||\mathcal{HC}[\mathbf{Z}^m_L]||_F$. In $\mathcal{M}(\cdot)$, we replace the original signal including both DC and HC components with the constant mask signal with the DC component only, so we can assume $||\mathcal{HC}[\mathbf{Z}^m]||_F<||\mathcal{HC}[\mathbf{Z}]||_F$. Then, we obtain $||\mathcal{HC}[\mathbf{Z}^m_L]||_F<||\mathcal{HC}[\mathbf{Z}^m]||_F<||\mathcal{HC}[\mathbf{Z}]||_F$. Obviously, the learning objective of MAE is pushing $||\mathcal{HC}[\mathbf{Z}^m_L]||_F\approx||\mathcal{HC}[\mathbf{Z}]||_F$ and that would make the $||\mathcal{HC}[\mathbf{Z}^m_L]||$ closer to its upper bound (\ie $||\mathcal{HC}[\mathbf{Z}]||_F$) during MAE training, which means the smoothing rate $\tau$ in Eq.~\ref{eq: smoothing rate} would be pushed towards greater implicitly to avoid the decay of high-frequency information. As a comparison, the model trained via simple classification target can still finish the task using totally identical token representations with DC component only. The reason is that the high-frequency information decay is not regularized by training objective.

The next question is whether $\tau$ can be trained towards greater. The answer is yes. First, $\tau$ is positive related with $||\mathbf{W}_V||_2$. At the same time, $\alpha$ in Eq.~\ref{eq: tau define} has an upper bound in Eq.~\ref{eq: alpha define}.
When model tends to provide more upside potential for $\alpha$, it has to increase $||\mathbf{W}_Q \mathbf{W}_K^T||_2$ or at least ensure $||\mathbf{W}_Q \mathbf{W}_K^T||_2$ would not decay during training.
Formally, let $n>2$, we can easily find $\frac{\partial\tau}{\partial \alpha}>0$ and $\frac{\partial\tau}{\partial ||\mathbf{W}_V||_2}>0$. According to Eq.~\ref{eq: alpha define}, we can increase the upper-bound of $\alpha$ via increasing $||\mathbf{W}_Q \mathbf{W}_K^T||_2$. Therefore, MAE can implicitly amplify the smoothing rate by increasing the value of $||\mathbf{W}_Q \mathbf{W}_K^T||_2$ and $||\mathbf{W}_V||_2$, and then alleviate the over-smoothing of Transformer.

\subsection{Quantitative Verification}\label{sec: quantitative verification}

We conduct quantitative verification to check our reasoning that MAE can implicitly maximize the smoothing rate. We initialize two large scale Transformer models (ViT-L/16 and MAE-L/16) with the same initializer but train them with ViT training objective and MAE pre-training objective, respectively. If our reasoning is correct, we can expect the model trained with MAE objective has greater $||\mathbf{W}_Q \mathbf{W}_K^T||_2$ and $||\mathbf{W}_V||_2$ than model trained with simple classification objective.

\begin{figure}[t]
\centering
\subfigure[Quantitative verification of $\beta_{QK}$ and $\beta_{V}$]{\label{fig: beta-qk-v}\includegraphics[width=0.42\textwidth]{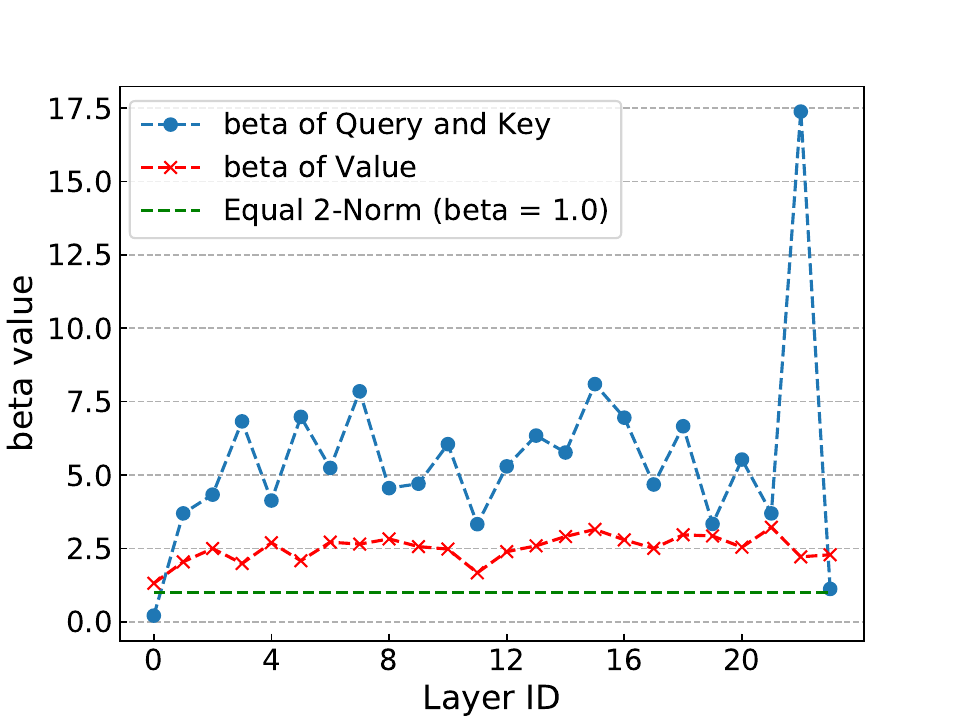}}\vspace{-0.4cm}
\subfigure[Quantitative verification of combining $\beta_{QK}$ and $\beta_{V}$]{\label{fig: beta-qk-v-combined}\includegraphics[width=0.42\textwidth]{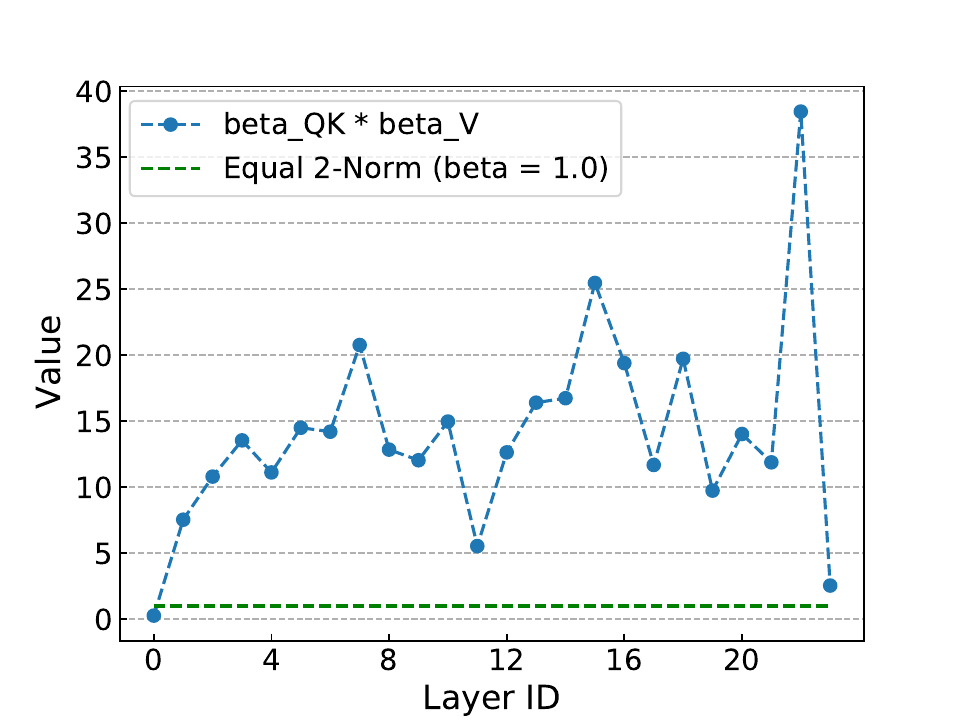}}\vspace{-0.35cm}
\caption{Quantitative verification of our theoretical reasoning. $beta>1.0$ means the MAE is alleviating the over-smoothing issue at this layer.}
\vspace{-0.4cm}
\label{fig: beta}
\end{figure}

We define $\beta_{QK}=\frac{||\mathbf{W}^\mathrm{MAE}_Q {\mathbf{W}^\mathrm{MAE}_K}^T||_2}{||\mathbf{W}^\mathrm{ViT}_Q {\mathbf{W}^\mathrm{ViT}_K}^T||_2}$ and $\beta_{V}=\frac{||\mathbf{W}^\mathrm{MAE}_V||_2}{||\mathbf{W}^\mathrm{ViT}_V ||_2}$. 
$\beta>1$ means MAE tends to obtain larger smoothing rate than simple supervised learning. In that case, the MAE training objective is alleviating over-smoothing implicitly.
We visualize the $\beta$ value of different layers. The comparison is shown in Figure~\ref{fig: beta}. We found that, even if we initialize ViT and MAE with the same initializer, both $\beta_{QK}$ and $\beta_{V}$ are significantly greater than 1.0 (green line) for most layers after training, which matches well with our expectation.

\subsection{Over-smoothing on Different Objectives}

\begin{figure}[t]
\centering
\subfigure[CLIP patch-pair cosine similarity]{\label{fig:clip-over-smoothing}\includegraphics[width=0.42\textwidth]{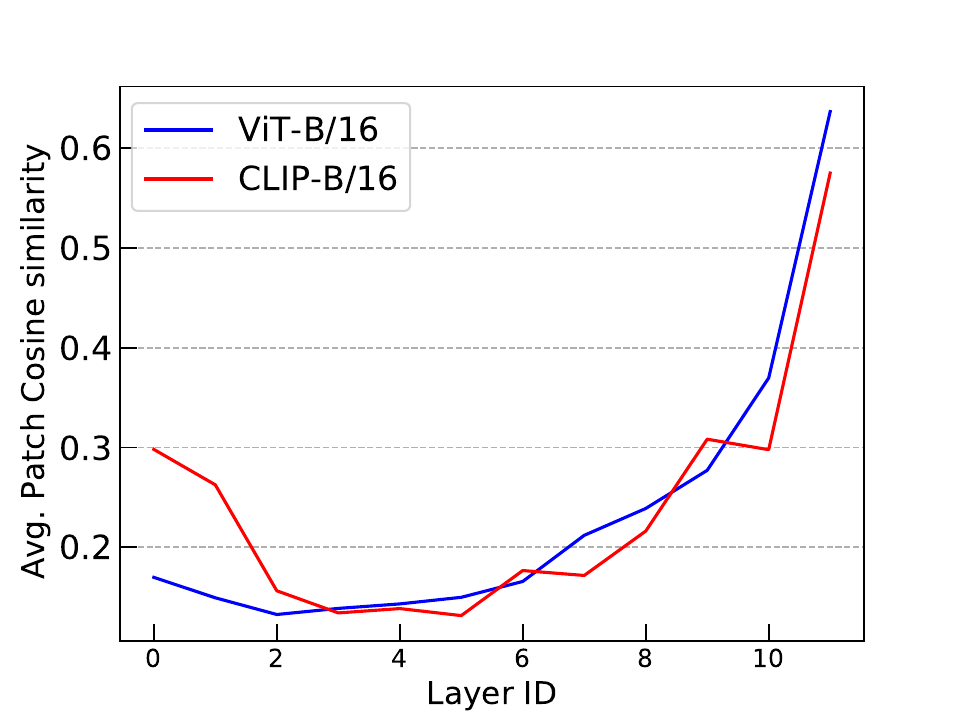}}\vspace{-0.4cm}
\subfigure[OPT token-pair cosine similarity]{\label{fig:opt-over-smoothing}\includegraphics[width=0.42\textwidth]{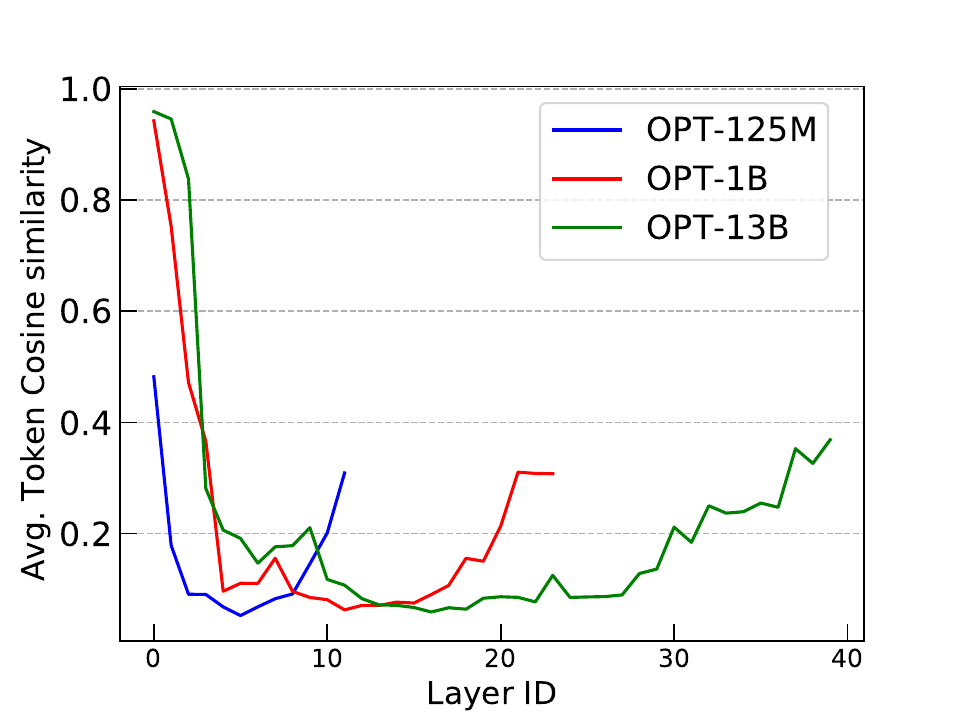}}\vspace{-0.35cm}
\caption{Over-smoothing analysis of CLIP and OPT.}
\label{fig:clip-opt-over-smoothing}
\vspace{-0.4cm}
\end{figure}

Based on the analysis presented above, we can conclude that token-level training objectives, such as next-token prediction in Language Modeling, exhibit a less severe over-smoothing issue. On the other hand, sequence-level objectives, like contrastive image pre-training, are more prone to over-smoothing. To validate this conclusion, we conducted cosine similarity experiments using CLIP~\citep{radford2021learning} and OPT~\cite{zhang2022opt}.
Figure~\ref{fig:clip-over-smoothing} presents the results of the CLIP model, demonstrating a similar over-smoothing behavior to Vanilla ViT (Vision Transformer). This observation aligns with our expectations. Furthermore, to investigate whether the over-smoothing issue can be mitigated by next token prediction, a widely employed LLM pre-training objective, we evaluated OPT and found that it effectively addresses over-smoothing. This finding is significant as it helps to elucidate why LLM models exhibit greater scalability compared to numerous vision models.

\section{Bamboo}\label{Bamboo}

\begin{table*}[t]
\caption{The configurations we used to scale transformer along depth. For a fair comparison, we keep a comparable computation cost with the original transformer configurations (\ie depth=12 for Base, depth=24 for Large).}
\centering
\small
\begin{tabular}{l llll llll}
\toprule 
Scale                  & \multicolumn{4}{l}{Base} & \multicolumn{4}{l}{Large} \\  \cmidrule(lr){2-5}\cmidrule(lr){6-9}
Depth             & 12   & 24   & 48   & 96  & 24    & 48   & 60   & 96  \\ 
Width             & 768  & 512  & 384  & 256 & 1024  & 768  & 640  & 512 \\
\#Attention Heads & 12   & 8    & 6    & 4   & 16    & 12   & 10   & 8   \\
FLOPs  & 1$\times$    & 0.9$\times$  & 1$\times$    & 0.9$\times$ & 1$\times$    & 1.1$\times$  & 1$\times$    & 1$\times$  \\ \bottomrule
\end{tabular}
\vspace{-0.4cm}
\label{tbl:config-explore}
\end{table*}

\begin{figure}[t]
\centering
\subfigure[Scaling supervised ViT along depth]{\label{fig:sweet-base}\includegraphics[width=0.42\textwidth]{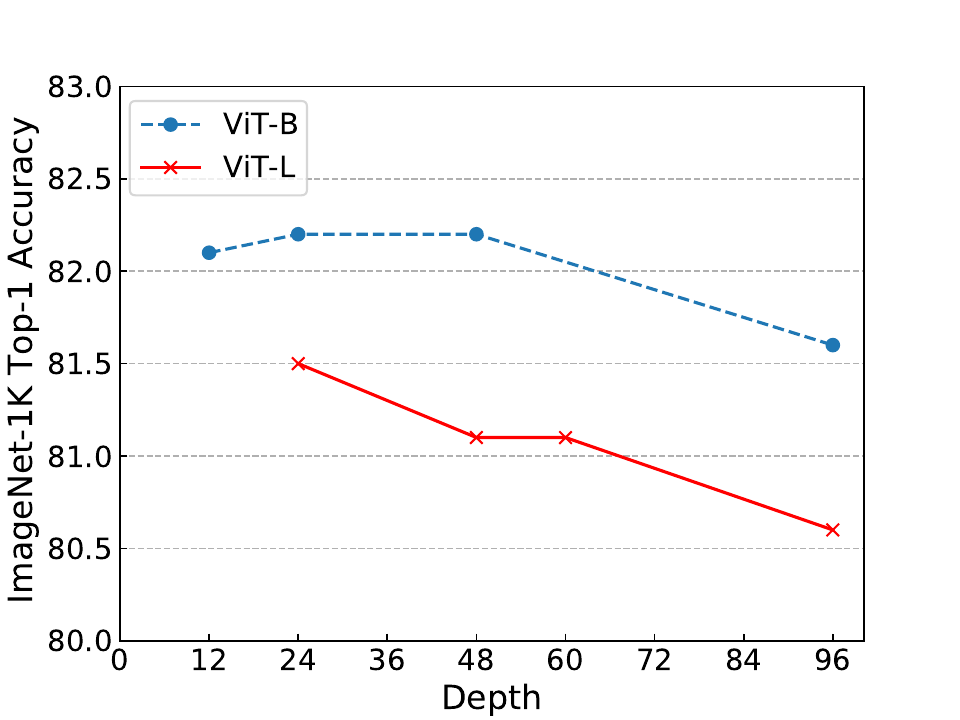}}\vspace{-0.35cm}

\subfigure[Scaling MAE along depth]{\label{fig:sweet-large}\includegraphics[width=0.42\textwidth]{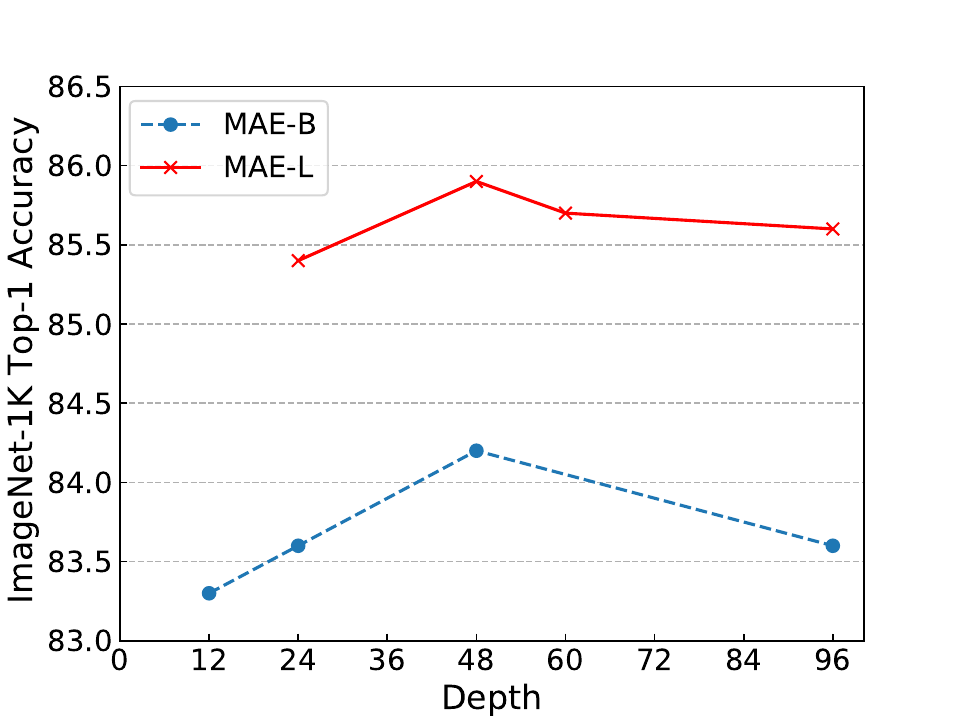}}\vspace{-0.35cm}
\caption{Scaling ViT and MAE with comparable number of parameters and computation cost. For deeper model, we compress the width to ensure a fair comparison.}
\label{fig:sweet}
\vspace{-0.4cm}
\end{figure}

Our analysis above shows that the masked autoencoder helps to alleviate the over-smoothing issue, which is a main challenge in scaling transformer along depth~\cite{zhou2021deepvit,dong2021attention}. To realize this potential, we suggest that we may obtain better performance by adapting deeper and narrower model configuration when training with MAE objective. We name this idea as Bamboo in this paper due to the shape of new transformer configuration.
Certainly, infinite deeper and narrower configurations do not always improve performance, as there are a few other reasons hindering scaling transformer along depth. In this section, we devote to find a sweet point following the Bamboo idea, to achieve effective masked autoencoder by experiments.



To validate our reasoning above, we conduct experiments on training transformer on ILSVRC-2012 ImageNet~\citep{deng2009imagenet} (ImageNet-1K), and report the top-1 accuracy. We use the original ViT trained by supervised learning and MAE pre-trained masked image modeling as the test platform. For a fair comparison with the original transformer configurations, when we scale transformer to deeper, we also reduce the width to keep comparable number of trainable parameters and computation cost. The configurations we used are summarized in Table~\ref{tbl:config-explore}.

For supervised learning models (\ie ViT), we train base scale models for 300 epochs and large scale models for 200 epochs following \citet{he2021masked}. We use RandAugment~\citep{cubuk2020randaugment}, drop path~\citep{huang2016deep}, mixup, cutmix~\citep{yun2019cutmix}, label smoothing~\citep{szegedy2016rethinking} for data augmentation.  Detailed hyper-parameters are summarized in Appendix~\ref{appendix:hyper-parameter}. For the masked autoencoder, we pre-train the base scale models for 1600 epochs and fine-tune for 100 epochs. For large models, we pre-train for 800 epochs and fine-tune for 50 epochs.


The results of scaling to deeper transformers are summarized in Figure~\ref{fig:sweet}. For both the base-scale and large-scale models, we report the top-1 accuracy on ImageNet-1K dataset. We can observe the models that are trained by supervised image classification directly (\ie ViT-B and ViT-L) cannot improve accuracy when we use  deeper architectures. For ViT-B, the top-1 accuracy on ImageNet-1K remains the same when the depth is smaller than 50, but after that, there is a significant drop on accuracy. For ViT-L, the accuracy decreases even faster than the ViT-B. There is a significant drop on accuracy  when we use the narrower and deeper models. However, for masked autoencoders, we observe totally different patterns. Even if we keep the comparable trainable parameters and computation cost, with such a simple modification, the masked autoencoders gain significant improvements. More importantly, when we scale to 48 layers, both 
MAE-B and MAE-L reach sweet spots. When scaling to 96 layers, we observe the training is unstable compared with shallower models. This result matches well with our observation in Figure~\ref{fig:over-smoothing-cos}. At around 40th layer, the over-smoothing issue starts to happen slightly in masked autoencoder, which is much later than the ViT model. We suggest another reason of the unstable deep model training is the too-large model updates~\citep{wang2022deepnet}. In this work, we focus on the over-smoothing issue of deep transformer training instead. The large model updates in deeper layers are out-of-scope. We leave that as our future work.

\begin{table*}[t]
\caption{Re-designed configurations under Bamboo idea. The computation cost denotes the FLOPs compared with the original configuration.}
\centering
\small
\begin{tabular}{l ll ll ll}
\toprule 
Scale                  & \multicolumn{2}{l}{Base} & \multicolumn{2}{l}{Large} & \multicolumn{2}{l}{Huge} \\  \cmidrule(lr){2-3} \cmidrule(lr){4-5}\cmidrule(lr){6-7}
                 & Original & Bamboo & Original & Bamboo & Original & Bamboo  \\ \midrule
Depth             & 12   & 48   & 24   & 48    & 32   & 64    \\ 
Width             & 768  & 384  & 1024  & 768 & 1280  & 896 \\
\#Attention Heads & 12   & 6    & 16    & 12   & 16    & 14    \\
Computation cost  & 1$\times$    & 1$\times$  & 1$\times$    & 1.1$\times$ & 1$\times$    & 1$\times$   \\ \bottomrule
\end{tabular}
\label{tbl:bamboo}
\vspace{-0.4cm}
\end{table*}

According to the experimental results above, we find that masked autoencoder can indeed scale transformer well along depth. Even if we keep comparable trainable parameters and computation cost with the original transformer, the model achieves better accuracy. Another observation is, both transformer-base and transformer-large reach their sweet spots at around 50 layers. We thus recommend a new set of transformer configurations in Table~\ref{tbl:bamboo} following our Bamboo idea, which are deeper and narrower than the original transformer configurations.

\begin{table*}[t]
\caption{Top-1 accuracy on ImageNet-1K. We report two versions of ViT training from scratch. The first one is from original ViT paper~\citep{dosovitskiy2020image}, and the second one is from \citet{he2021masked}'s re-implementation with strong data augmentation. For MAE-B, we reproduce the results by running the official code and obtain a slightly different result (denoted by 83.3*). The original result is 83.6.}
\centering
\small
\begin{tabular}{l llll}

\toprule 
Method             & Pre-train Data    & Base & Large & Huge \\ \midrule
ViT from scratch~\citep{dosovitskiy2020image}            & - & 77.9 & 76.5 & - \\
DeepViT~\citep{zhou2021deepvit}             & - & 80.9 & - & - \\ 
DeiT~\citep{touvron2021training}             & - & 81.8 & - & - \\
ViT from scratch~\citep{he2021masked}            & - & 82.1 & 81.5 & 80.9  \\ \midrule
DINO~\citep{caron2021emerging}             & IN1K & 82.8 & - & -  \\
MoCo v3~\citep{chen2021empirical}             & IN1K & 83.2 & 84.1 & -  \\ \midrule
BEiT~\citep{bao2021beit}             & IN1K + DALL-E & 83.2 & 85.2 & -  \\
MaskFeat~\citep{wei2021masked}             & IN1K & 84.0 & 85.7 & -  \\
IBOT~\citep{zhou2021ibot}             & IN1K & 84.0 & 84.8 & -  \\ \midrule
MAE~\citep{he2021masked} (Direct baseline)           & IN1K & 83.3* & 85.9 & 86.9 \\
Bamboo (Ours)             & IN1K & \textbf{84.2} & \textbf{86.3} & \textbf{87.1} \\\bottomrule
\end{tabular}
\label{tbl:exp-cv-sota}
\vspace{-0.4cm}
\end{table*}

\section{Evaluation on Vision Tasks}\label{Evaluation Vision}

\subsection{Settings} 

We further evaluate our deeper and narrower configurations on vision task. We train different models with more training epochs and compare with SoTA vision models. We conduct experiments on ImageNet-1K and compare with recent supervised vision models \eg DeepViT~\citep{zhou2021deepvit} and DeiT~\citep{touvron2021training},  and self-supervised vision models \eg DINO~\citep{caron2021emerging}, MoCo v3~\citep{chen2021empirical}, BEiT~\citep{bao2021beit} and MAE~\citep{he2021masked}. Compared with the MAE, \textit{the only difference is the configurations}. That is, MAE uses original transformer configurations and we use our Bamboo configurations. The experiments are to verify that such a simple modification can improve model effectiveness and show the potential of more reasonable configurations. 

We evaluate the models on three different scales, \ie base, large, and huge. The data augmentation setting is exactly the same as MAE for a fair comparison. Again, the only difference is that we use the Bamboo configurations instead of the original transformer configurations. During fine-tuning, we use the same script with training ViT from scratch. We fine-tune base models for 100 epochs. For large and huge models, we fine-tune them for 50 epochs following existing work~\citep{bao2021beit, he2021masked}.

\subsection{Results}

Results on ImageNet-1K are reported in Table~\ref{tbl:exp-cv-sota}. For training ViT from scratch, we report the original results~\citep{dosovitskiy2020image} and the results with strong data augmentation~\citep{he2021masked}. A few recent work~\citep{zhou2021deepvit,touvron2021training} focusing on training ViT from scratch on ImageNet-1K are also included. In general, we can find the models pre-trained by self-supervised learning (\eg DINO, MoCO v3, MAE) perform much better than training from scratch. If we only consider the self-supervised learning approaches, masked image modeling-based methods (\eg BEiT, MaskFeat, IBOT, MAE) outperform the contrastive learning-based methods (\eg DINO, MoCo v3) significantly, especially on larger scale. 

Since we are focusing on the scalability of training transformer with masked autoencoder, we choose MAE as our direct baseline. We train MAE with Bamboo configurations and report the results in Table~\ref{tbl:exp-cv-sota}. Observe that our Bamboo achieves the best top-1 accuracy on all scales. On the base scale, Bamboo achieves state-of-the-art performance, 84.2 top-1 accuracy, which is 0.9 (0.6) points higher than MAE-B. When we scale the model up to large scale and huge scale, Bamboo remains the best performer and achieves 86.3 and 87.1 top-1 accuracy respectively. Note that, compared to other scales, the improvement is not so significant on the huge scale. One reason is, the original huge configuration has been deep (\ie 32 layers), which is close to the sweet point. Similarly, this can also explain why the configurations designed under Bamboo can improve MAE-B significantly. Since the real run time may be influenced by many other factors (\eg GPU or TPU utilization), we report the real throughput in the Appendix~\ref{appendix:throughput}.

\begin{table*}[t]
\caption{Results of fine-tuning on GLUE benchmark. }
\vspace{-0.5cm}
\label{nlp:table:main}
\small
\begin{center}
\begin{tabular}{l llllllll l}

\toprule
Method                   & CoLA & MNLI & MRPC & QNLI & QQP & RTE & SST-2  & STS-B & Avg \\ \midrule
BERT-B   & 59.6 & 83.7 & 88.0 & 90.4 & 88.8 &  68.6  & 91.5 & 89.4 & 82.5 \\ 
Bamboo-B   & \textbf{60.5} & \textbf{84.3} & \textbf{88.4} & \textbf{90.5} & \textbf{89.0} & \textbf{70.0} &  \textbf{92.2}  & \textbf{89.5} & \textbf{83.1} \\  \midrule
BERT-L   & 60.9 & 86.2 & 89.3 & 92.3 & \textbf{89.6} &  73.1  & 92.5 & 90.4 & 84.3 \\
Bamboo-L   & \textbf{62.9} & \textbf{87.1} & \textbf{89.8} & \textbf{92.4} & 89.4 &  \textbf{77.3}  & \textbf{93.8} & \textbf{90.6} & \textbf{85.4} \\ 
\bottomrule
\end{tabular}
\vspace{-0.4cm}
\end{center}
\end{table*}

\section{Evaluation on Language Tasks}\label{Evaluation Language}


We further evaluate our Bamboo configurations on language tasks. We select BERT~\citep{devlin2018bert} as the platform to evaluate our Bamboo configuration because it is widely used on many language tasks. Note that BERT is a post-layer normalization transformer model. Better performance on BERT means our design can generalize to other architectures. We follow the BERT paper to use Wikipedia and bookscorpus to pre-train. During pre-training, we use LAMB optimizer and set batch size and learning rate as 4096 and 1.76e-3, respectively, following~\citet{you2019large}.


During fine-tuning, we conduct experiments on General Language Understanding Evaluation (GLUE) benchmark.
The GLUE benchmark~\citep{wang-etal-2018-glue} is widely used in natural language understanding tasks, which include 8 tasks, \ie CoLA, MNLI, MRPC, QNLI, QQP, RTE, SST-2 and STS-B. We set the learning rate as 1e-5 or 2e-5. Batch size is fixed as 32. For small datasets, \ie CoLA, MRPC, RTE and STS-B, we fine-tune for 10 epochs. For larger datasets, \ie MNLI, QNLI, QQP and SST-2, we fine-tune for 3 epochs. Matthew’s correlation is used as metric for CoLA. For MNLI, we report the average accuracy on MNLI-m and MNLI-mm. QNLI and RTE also adapt the accuracy as metric. The results on MRPC and QQP are reported with the average of F1 and accuracy. We use Spearman correlation on STS-B. We run the code 5 times and report the median for fine-tuning.


The results on language tasks are reported in Table~\ref{nlp:table:main}. Under the same pre-training and fine-tuning settings, models with Bamboo configurations outperform BERT significantly. Bamboo-L achieves the best performance in Table~\ref{nlp:table:main}. Compared with BERT-L, our Bamboo-L wins on 7 out of 8 datasets, and surpassed BERT-L by 1.1 points on average. It is also notable Bamboo can outperform baselines on both large datasets (\eg MNLI) and small datasets (\eg CoLA).

\section{Discussion}\label{Discussion}

\textbf{Compared with simply scaling along depth}, this work maintains a comparable computation cost and the number of trainable parameters. When scaling along depth, we also make the deep transformer narrower. Under such setting, the deeper and narrower configurations re-designed under the Bamboo idea can still outperform the baseline configurations, suggesting that we should consider narrower and deeper transformer when training by masked autoencoder. One related work is~\citep{tay2021scale}, an empirical study of practical scaling of transformer, which has a similar observation, deeper and narrower can improve accuracy. 
On this basis, our finding is consistent with that in~\citep{tay2021scale}. 
However, we are not simply comparing different configurations~\citep{tay2021scale} or training objectives~\citep{voita2019bottom}. We are actually bridging the gap and study the relation between configurations and the training objective. Another related work~\citep{levine2020depth} investigates the optimal depth-to-width ratio for different scales. However, they do not tackle the impact of training objectives.

\textbf{Compared with brute-force hyper-parameter tuning}, we provide both theoretical reasoning and quantitative analysis to justify our insight, \ie masked autoencoder alleviates over-smoothing issue in transformer. Motivated by this, we suggest using deeper and narrower model for masked autoencoder. There is no guarantee to ensure the configurations are always optimal. We believe it is impossible in deep learning to know the optimal configurations before experiments. However, we argue that the masked autoencoder gets more benefits from deeper configurations. Our insight may instruct future work to consider configurations according to the training objectives.

\textbf{Instead of proposing a new approach or a new set of configuration,} this work focuses on an existing but neglected problem. After revisiting, we find configurations should be re-designed for different training objectives. We highlight this is important as following the conventional configuration is not the real ``fair'' comparison. If we keep do this in the future, we will always pick the training objective that wins the ``configuration lottery'', and that would miss some real effective objective with great potential. The analysis sections in this paper can be seen as an example of how to design configurations for different objectives. After simply re-designing a set of configurations for masked autoencoder, we can see a significant improvement on both vision and language tasks. On vision tasks, we even achieve SoTA top-1 accuracy on ImageNet. However, note that we highlight our main contribution is an insight instead of a SoTA model, and it is orthogonal to the future transformer modifications.


\section{Conclusion}\label{Conclusion}

In this work, we first study the relationship between transformer configuration and the training objective. Compared with supervised learning, training transformer with MAE can alleviate over-smoothing. We then explore the reason behind this finding through theoretical reasoning and quantitative verification via Fourier domain analysis. Under this insight, we rethink the widely used configurations in vision and language tasks, and suggest deeper and narrower configurations when training with MAE. To further verify the effectiveness of our configuration, we conduct comprehensive experiments on both large-scale vision and language tasks and achieve significant improvement with such a simple modification. More importantly, we argue that using a configuration for a fair comparison may not be really fair. That may underestimate the potential of a new training objective who does not win the ``configuration lottery''. We suggest analyzing the inductive bias of each objective and sweeping the configuration following the analysis. 

\section*{Acknowledgement}
This work is being sponsored by Huawei Noah's Ark Grant.

\bibliography{example_paper}
\bibliographystyle{icml2023}

\newpage
\appendix
\onecolumn

\section{Fine-tuning Hyper-parameters}\label{appendix:hyper-parameter}

\begin{table}[ht]
\caption{Hyper-parameters on ImageNet fine-tuning}
\label{tbl-hyper-parameter-pre-train}
\vskip 0.15in
\begin{center}
\begin{tabular}{l|ccc}
\toprule
Parameter                  & Base  & Large & Huge   \\ \midrule
Epoch                     & 100     & 50  & 50    \\
Warmup Epochs             &      & 5    &        \\
Batch Size                &    & 1024  &       \\
Learning rate             &    & 2e-3  &    \\
Layer-wise learning rate decay & 0.65  & 0.75 & 0.75  \\
Weight Decay              &     & 0.05 &   \\
DropPath                  & 0.1   & 0.2 & 0.2    \\ 
Label smoothing           &     & 0.1  &       \\
Erasing prob.               &     & 0.25  &     \\
RandAug               &   & 9/0.5 &       \\
Mixup prob.               &    & 0.8 &       \\
Cutmix prob.               &    & 1.0 &       \\ \bottomrule
\end{tabular}
\end{center}
\vskip -0.1in
\end{table}

\section{Throughput Comparison}\label{appendix:throughput}

\begin{table*}[ht]
\caption{Throughput comparison of re-designed configurations under Bamboo idea. The throughput here means the image precessed per second by one TPU core. is measured during MAE pre-training. For base and large-level models, we use 128 TPUv3 cores in parallel. For the huge models, we use 256 TPUv3 cores.}
\vspace{5pt}
\centering
\begin{tabular}{l ll ll ll}
\toprule 
Scale                  & \multicolumn{2}{l}{Base} & \multicolumn{2}{l}{Large} & \multicolumn{2}{l}{Huge} \\  \cmidrule(lr){2-3} \cmidrule(lr){4-5}\cmidrule(lr){6-7}
                 & Original & Bamboo & Original & Bamboo & Original & Bamboo  \\ \midrule
Computation Cost  & 1$\times$    & 1$\times$  & 1$\times$    & 1.1$\times$ & 1$\times$    & 1$\times$   \\ 
Throughput  & 1$\times$    & 0.9$\times$  & 1$\times$    & 0.9$\times$ & 1$\times$    & 0.9$\times$   \\ \bottomrule
\end{tabular}
\vspace{-0.3cm}
\label{tbl:throughput}
\end{table*}

From Table~\ref{tbl:throughput}, we can see the deeper configurations are slightly slower, although they have comparable computation cost. However, we highlight that is fine. There is a trade-off instead of a limitation when using narrow configuration. During inference, we can actually do the inference layer by layer and only load one transformer layer into memory. After using, we can offload the layer and load the next one in. Such a design can be found in Figure 5 of SE-MoE paper~\citep{shen2022se}. Since our single layer is narrower and includes fewer parameters, our model is more memory-efficient during inference.   

\newpage

\section{More Figures}

\begin{figure}[h]
\centering
\begin{tabular}{cc}
    \subfigure[ImageNet Top-1 Accuracy improvement when scaling ViT-B and MAE-B across depth.]{\label{fig:base-acc-improve}\includegraphics[width=0.45\textwidth]{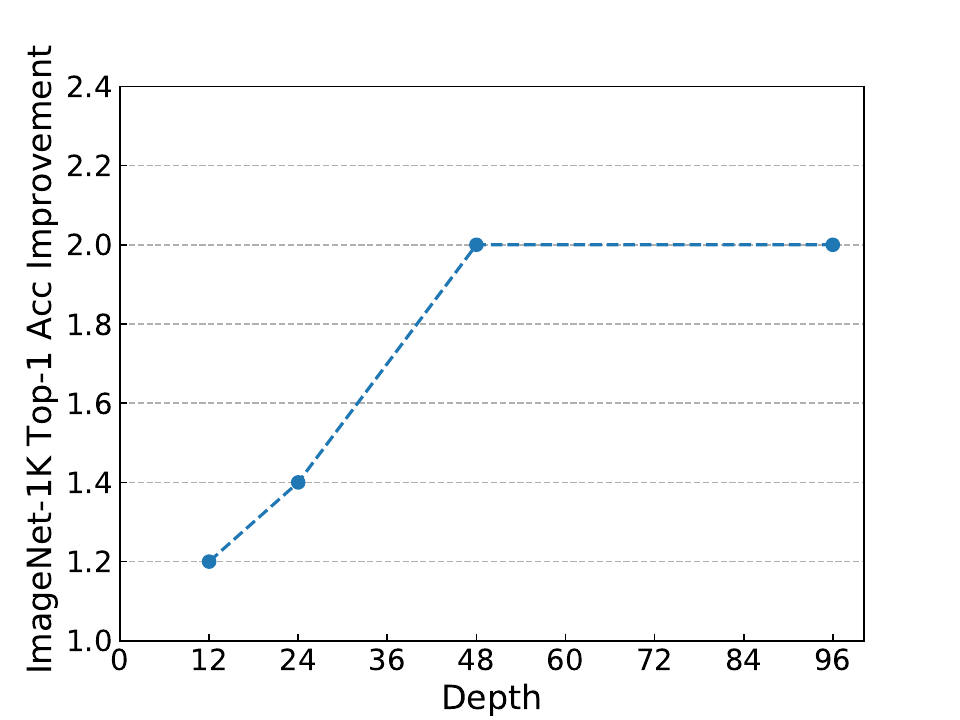}} &
    \subfigure[Average standard deviation gap when scaling ViT-B and MAE-B across depth.]{\label{fig:base-smooth-gap}\includegraphics[width=0.45\textwidth]{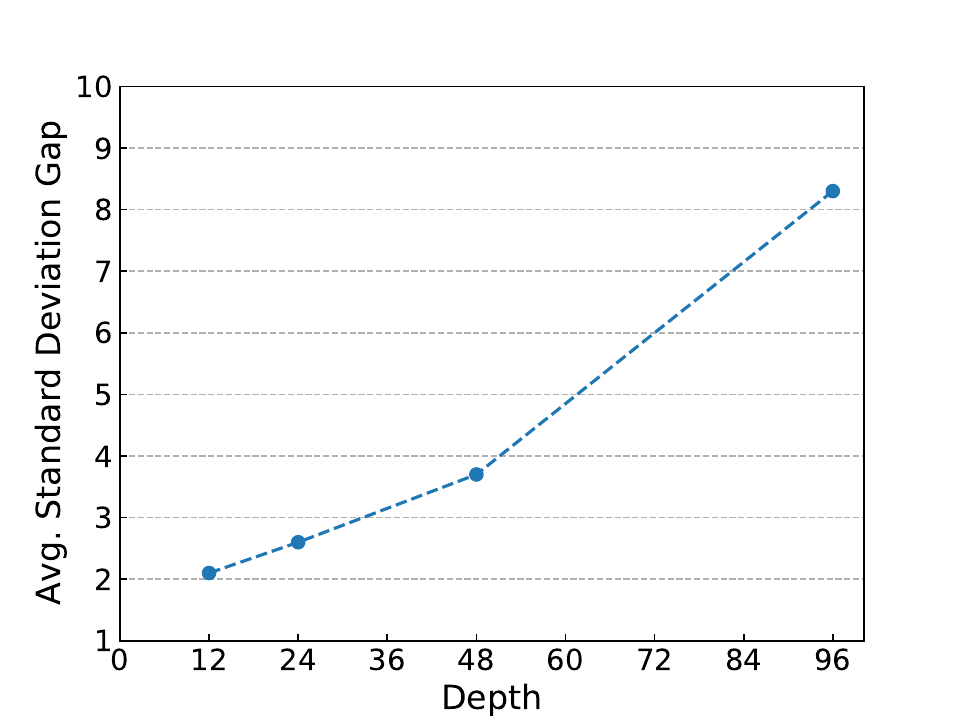}} \\
    \subfigure[ImageNet Top-1 Accuracy improvement when scaling ViT-L and MAE-L across depth.]{\label{fig:large-acc-improve}\includegraphics[width=0.45\textwidth]{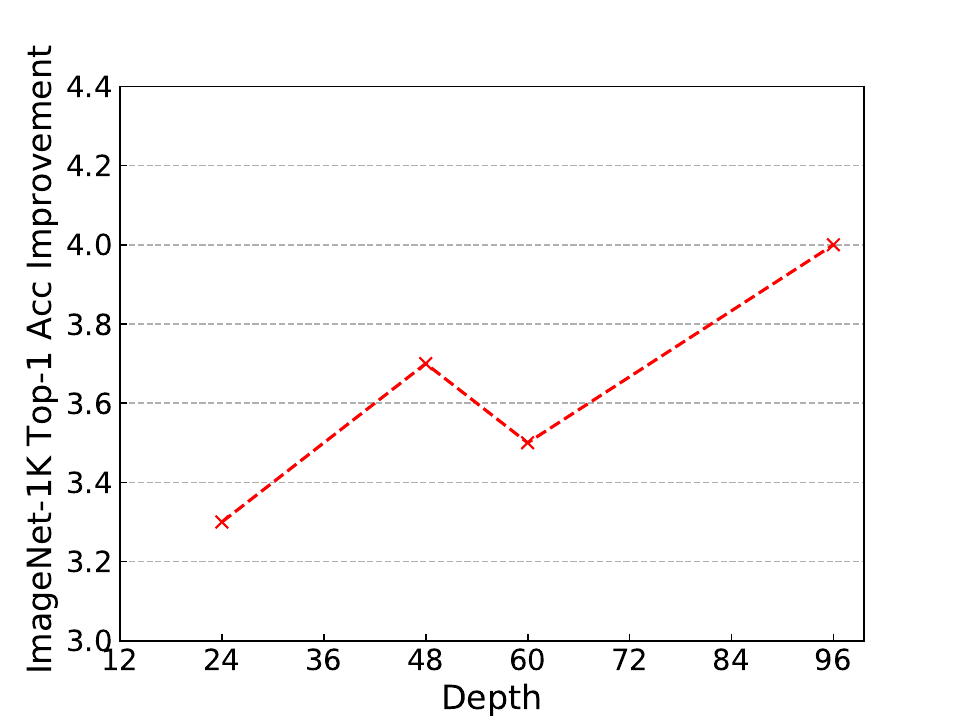}} &
    \subfigure[Average standard deviation gap when scaling ViT-L and MAE-L across depth.]{\label{fig:large-smooth-gap}\includegraphics[width=0.45\textwidth]{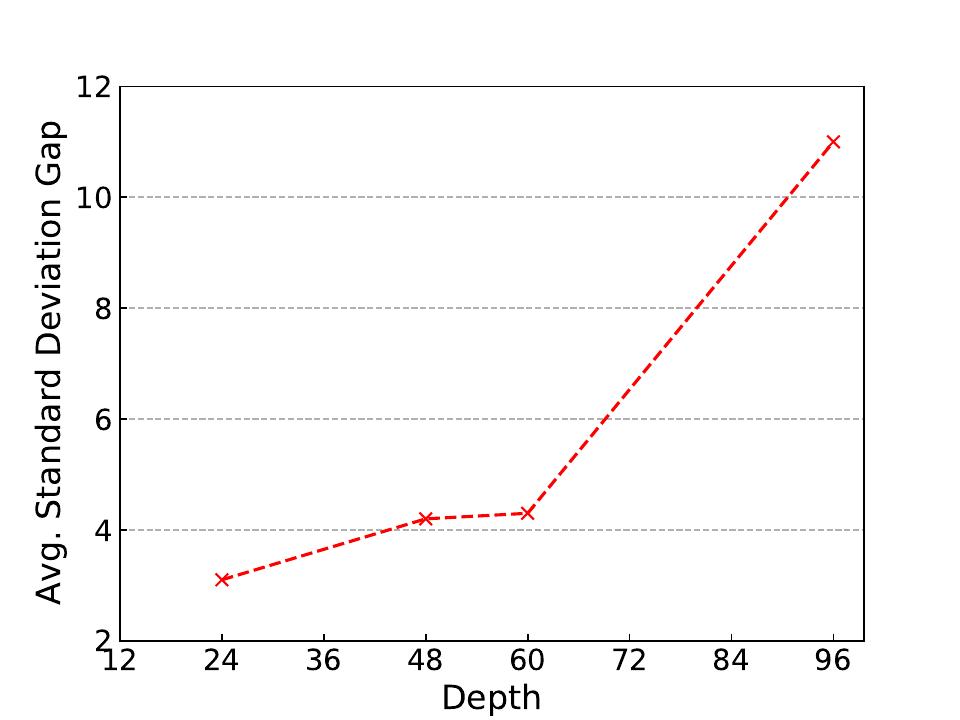}} \\
\end{tabular}
\caption{We compare the ImageNet Top-1 accuracy and average standard deviation of ViT and MAE models with different configurations. A larger average standard deviation gap means that MAE training objective alleviates more over-smoothing issue.}
\label{fig:}
\end{figure}






\end{document}